\documentclass[conference]{IEEEtran}
\ifCLASSINFOpdf
\else
\fi

\usepackage{multirow}
\usepackage{graphicx}
\usepackage{subcaption}
\usepackage{comment}
\usepackage{subscript}

\begin{document}
%
\title{Explainable Landscape-Aware Optimization Performance Prediction}

\author{\IEEEauthorblockN{Risto Trajanov}
\IEEEauthorblockA{Faculty of Computer Science and Engineering\\ Ss. Cyril and Methodius, University\\
Skopje, North Macedonia\\
risto.trajanov@gmail.com}
\and
\IEEEauthorblockN{Stefan Dimeski}
\IEEEauthorblockA{Faculty of Computer Science and Engineering\\ Ss. Cyril and Methodius, University\\
Skopje, North Macedonia\\
stefan.dimeski.1@students.finki.ukim.mk}
\and
\IEEEauthorblockN{Martin Popovski}
\IEEEauthorblockA{Faculty of Computer Science and Engineering\\ Ss. Cyril and Methodius, University\\
Skopje, North Macedonia\\
martin.popovski@students.finki.ukim.mk}
\and
\IEEEauthorblockN{Peter Koro\v{s}ec}
\IEEEauthorblockA{Computer Systems Department\\ Jo\v{z}ef Stefan Institute\\
Ljubljana, Slovenia\\
peter.korosec@ijs.si}
\and
\IEEEauthorblockN{Tome Eftimov}
\IEEEauthorblockA{Computer Systems Department\\ Jo\v{z}ef Stefan Institute\\
Ljubljana, Slovenia\\
tome.eftimov@ijs.si}}

\maketitle

\begin{abstract}
Efficient solving of an unseen optimization problem is related to appropriate selection of an optimization algorithm and its hyper-parameters. For this purpose, automated algorithm performance prediction should be performed that in most commonly-applied practices involves training a supervised ML algorithm using a set of problem landscape features.  
However, the main issue of training such models is their limited explainability since they only provide information about the joint impact of the set of landscape features to the end prediction results. In this study, we are investigating explainable landscape-aware regression models where the contribution of each landscape feature to the prediction of the optimization algorithm performance is estimated on a global and local level. The global level provides information about the impact of the feature across all benchmark problems' instances, while the local level provides information about the impact on a specific problem instance. The experimental results are obtained using the COCO benchmark problems and three differently configured modular CMA-ESs. The results show a proof of concept that different set of features are important for different problem instances, which indicates that further personalization of the landscape space is required when training an automated algorithm performance prediction model.
\end{abstract}

\IEEEpeerreviewmaketitle

\section{Introduction}
\label{sec:introduction}

Nowadays, many real world scenarios involve optimization problems that should be solved. To solve them, the most appropriate optimization algorithm should be selected and its hyper-parameters should be set. This is a well-known problem of algorithm selection (AS)~\cite{bischl2012algorithm,kerschke2019automated,vermetten2020towards} and algorithm configuration (AC)~\cite{stutzle2015automated}, respectively. The prerequisite to achieve this is performing automated algorithm performance prediction. This involves learning a predictive model that is based on the properties of an optimization problem and is able to predict the optimization algorithm’s performance.

Landscape analysis is used to describe the properties of the problem instances, since it allows features extraction by using mathematical and statistical methods. For this purpose, exploratory landscape analysis (ELA)~\cite{mersmann2011exploratory,munoz2014exploratory} is used, which calculates low-level features from a sample of observations for a given problem instance. With regard to the computational cost associated with the ELA features calculation, they can be split into cheap (only from one initial sample) or expensive (additional sampling is required during the calculation). The R package \texttt{flacco}~\cite{kerschke2017package,kerschke2019comprehensive} conveniently computes up to 343 different feature values that originate from 17 different features groups.

Recent studies for automated algorithm selection use ELA features to represent the optimization problems, which are further used to train regression models that can predict the algorithm performance~\cite{KerschkeT19,BelkhirDSS17,MunozSurvey15,DerbelLVAT19,jankovic2020landscape}. In all these studies, typically a single-target regression (STR) model is learned to predict an algorithm's performance across all problems included in the benchmark platform. When performances of multiple algorithms are predicted, then a separate STR model is trained for each algorithm. All these models share the same ELA features as the input data and differ only in the performance targets. One issue in training those models is the selection of the ELA features that will be used to describe the problems. Some of the studies used the cheap ELA features, and some of them select a subset of the previously mentioned features using classical ML feature selection techniques~\cite{KerschkeT19,jankovic2021towards,renau2021towards}. The presented results showed that the selection of the ELA features portfolio influences the end performance prediction, but the impact of each feature to the end prediction result is still treated in a black-box manner.

To go beyond the black-box interpretation of the influence of each ELA feature to the end performance prediction result, we investigate how a ML regression model is making its predictions by analyzing the importance of each feature. Even more, to reduce the time needed to train multiple STR models (i.e., for each algorithm a separate STR model), in this paper, we also investigate the multi-target regression (MTR) that instead of learning multiple STR models learns only one model that simultaneously can predict the performances of all algorithms involved in the study~\cite{spyromitros2016multi}. The benefit of learning such a model is that beside the exploration of the relationships that exist between the ELA features and the performance targets, it implicitly investigates the inter-relationships that exist only in the performance space (i.e., between the performances of all algorithms involved in the study). Further, for the same learning task, we analyze the importance of each ELA feature in STR and MTR scenario to estimate the useful information that each feature conveys.

\textbf{Structure of the Paper:} The paper is organized as follows:  Section~\ref{sec:related_work} presents the related work. In Section~\ref{sec:mtr}, the ML pipeline for automated algorithm performance prediction is presented. The results of the study together with a critical discussion of proposed approach are presented in Section~\ref{sec:results}. Finally, the conclusions and the directions for future work are presented in Section~\ref{sec:conclusions}.      

\textbf{Availability of Data and Code:} All project data and code is available at~\cite{data}.

\section{Related work}
\label{sec:related_work}

Recent studies focused on AS and AC explored ELA-based algorithm performance prediction~\cite{KerschkeT19,BelkhirDSS17,MunozSurvey15,DerbelLVAT19}. Kerschke and Trautmann~\cite{kerschke2019automated1} presented ML models for automated AS, based on the ELA problem representations, which predicted the best performing optimization algorithm. One of the investigated solutions is a regression approach, where STR model is trained on performances of different optimization algorithms, out of which the best solver is selected. To do this, they analyzed recursive partitioning and regression trees~\cite{strobl2009introduction}, kernel-based support vector machine~\cite{melki2017multi}, random forest~\cite{xi2018empirical}, and extreme gradient boosting~\cite{joly2019gradient} as ML regression models in combination with four classical feature selection techniques: greedy forward-backward selection~\cite{borboudakis2021extending}, greedy backward-forward selection~\cite{borboudakis2021extending}, and two variants of (10+5)- genetic algorithm~\cite{babatunde2014genetic}. 

Jankovic and Doerr~\cite{jankovic2020landscape} analyzed automated algorithm selection by combining two ELA-based regression models for selecting an algorithm that performs best within a given budget of function evaluations. For this purpose, random forest has been investigated with no hyper-parameter tuning in STR learning scenario to predict the target precision reached by an algorithm after a fixed number of function evaluations and its logarithmic value, separately. Further, a comprehensive study that analyzed the impact of the hyper-parameter tuning of tree-based ML regression models (e.g., decision trees, random forest, bagging, and boosting) in STR learning scenario to the landscape-aware performance prediction and algorithm selection has been presented~\cite{jankovic2021impact}. It showed that different ML models and different hyper-parameters are recommended to be used for different optimization algorithm prediction. Instead of learning one STR model that performs the best on average across all benchmark problems, Eftimov et al.~\cite{eftimov2021personalizing} investigated personalized performance regression models for each black-box optimization problem. For this purpose, ensemble of tree-based STR models have been selected for each benchmark optimization problem, which in general decreased the prediction errors within each problem.

In~\cite{jankovic2021towards}, trajectory-based ELA regression models have been developed to predict the CMA-ES target precision obtained after a fixed number function evaluations. It has been shown that classical feature selection techniques applied on the ELA features in combination with random forest with no hyper-parameter tuning did not lead to better performance results. 

All previously mentioned studies provide important steps in explaining automated performance prediction, however only the joint impact of the ELA features to the end prediction results can be analyzed using the classical feature selection techniques. To go beyond this and provide more in-depth explanation in automated algorithm performance prediction, we analyzed the global and local impact of each ELA feature on the performance prediction. Globally, the impact of each feature in the ML regression model is estimated using all benchmark problems, while locally the impact is analyzed for each benchmark problem separately.

\section{Explainable performance prediction}
\label{sec:mtr}

The automated algorithm performance prediction is investigated in two learning scenarios: single-target regression (STR) models, which are most commonly investigated in the literature, and multi-target (MTR) regression scenario that is to the best of our knowledge the first study that explores it. STR is a learning scenario where only one variable is predicted using the features portfolio data, while MTR is a learning scenario where at the same time multiple variables (i.e., at least two) are predicted using the same features portfolio data. 

Let us assume that each problem instance is described with $n$ ELA features $(x_1, x_2,\dots, x_n)$ to which a vector of $m$ performance targets are associated $(y_1, y_2,\dots, y_m)$. The targets can be the performance reached by algorithms when they are run on the same problem instance, or different transformations of the performance reached by a single algorithm (e.g., its original precision, logarithmic value of the precision, etc.). 

Having a training data set, STR learns $m$ STR models (i.e., one per each performance target), while MTR learns a model that for a new unseen problem instance can predict all $m$ targets at once. In a MTR learning scenario, the model learns only a single model by modelling the complex relationships between ELA features and performance targets, while at the same time implicitly exploring the inter-target relationships that exist between the $m$ performance targets. Since MTR is a well researched task in ML, different methods for learning such models already exist based on decision trees, random forests, neural networks, etc. 

To explain the regression models output in both STR and MTR learning scenarios and to understand how they make their predictions (i.e., explainable regression models), feature importance is analyzed considering Shapley values using the SHAP (SHapley Addictive exPlanations) method~\cite{molnar2020interpretable}. SHAP is a method to explain individual predictions~\cite{NIPS2017_7062}. The goal of SHAP is to explain the prediction of a single instance by computing the contribution of each feature to the prediction. The SHAP explanation method computes Shapley values from coalitional game theory. The feature values of a data instance act as players in a coalition. Shapley values tell us how to fairly distribute the ``payout" (i.e., the prediction) among the features. A player can be an individual feature value or a group of feature values. For example to explain an image, pixels can be grouped to super pixels and the prediction distributed among them. One innovation that SHAP brings to the table is that the Shapley value explanation is represented as an additive feature attribution method, a linear model.

The SHAP method provides two levels of explanations:
\begin{itemize}
    \item Global explanations - the SHAP values for each problem instance are combined to see the impact to the target variable (i.e., the impact across all problem instances involved in the learning process).
    \item Local explanations - each problem instance gets its own set of SHAP values that increases its transparency (i.e., it allows us to see how the importance of the features changes across different problem instances; each problem instance has its own set of Shapley values).
\end{itemize}

Next, we are going to briefly describe a use case that will be investigated for algorithm performance prediction. The multiple targets will present the target precision achieved by different algorithms. By using the same ELA features portfolio STR models will be investigated, one per each algorithm. Next, MTR will learn one model for performance prediction of different algorithms by exploring implicit relationships of the performances across the algorithms.

\section{Results and discussion}
\label{sec:results}
Next, the experimental setup is explained in more detail providing information about the landscape and performance data, followed by the results obtained for both use cases.
\subsection{Experimental setup}
\subsubsection{Landscape data} The COCO benchmark platform~\cite{hansen2020coco} is selected to represent the problem space. It consists of 24 single-objective continuous optimization problems. For our experiment, the problem dimension is set to $D=5$ and for each problem, the first 50 instances are included. Such experimental design leads to 1,200 problem instances, 50 per each problem. To calculate the landscape characteristics (i.e., ELA features) of the problem instances, the R package ``flacco" has been used~\cite{kerschke2017package}, from which 99 ELA features have been selected. The calculated ELA features are from the following groups: \emph{cm\_angle}, \emph{cm\_grad}, \emph{disp}, \emph{ela\_conv}, \emph{ela\_curv}, \emph{ela\_distr}, \emph{ela\_level}, \emph{ela\_local}, \emph{ela\_meta}, \emph{ic}, and \emph{nbc}. The selected ELA features are the cheap ones concerning their computation cost and do not have missing values for any problem instances. They have been calculated using the improved latin hypercube sampling (ILHS) method~\cite{xu2017improved} with $50D$ sample points. This process has been repeated 10 times for each problem instance, so each ELA feature value has been selected as the median value of its 10 repetitions. The median value has been selected since it is more statistically robust than the mean value.

\subsubsection{Performance data} As performance data three randomly selected configurations from modular CMA-ES have been investigated for showing the proof of concept. More information about the modular CMA-ES configurations can be found in~\cite{nobel2020}. In Table~\ref{tab:CMA_modules}, the selected modules configurations named CMA-ES$_1$, CMA-ES$_2$, and CMA-ES$_3$ are presented in more detail. Results for other 12 CMA-ES configurations are presented at our GitHub repository.

\begin{table}[ht]
\centering
\footnotesize
\caption{CMA-ES modules configurations.}
\label{tab:CMA_modules}
\begin{tabular}{l|ccc}
\hline
modules                & CMA-ES$_1$     & CMA-ES$_2$            & CMA-ES$_3$    \\
\hline
Active update                 & FALSE    & FALSE             & TRUE     \\
Elitism                & TRUE     & FALSE             & TRUE     \\
Orthogonal Sampling             & TRUE     & FALSE             & FALSE    \\
Sequential selection             & FALSE    & FALSE             & TRUE     \\
Threshold Convergence & TRUE     & TRUE              & FALSE    \\
Step Size Adaptation & tpa      & msr               & csa      \\
Mirrored Sampling              & mirrored & mirrored pairwise & mirrored \\
Quasi-Gaussian Sampling          & halton   & gaussian          & halton   \\
Recombination Weights        & default  & 1/mu              & default  \\
Restart Strategy         & BIPOP    & IPOP              & BIPOP \\   
\hline
\end{tabular}
\end{table}

Each CMA-ES configuration has been run 10 times on each problem instance in fixed budget scenario, where the budget is set to 50,000. In our case, the performance prediction is focused on the best reached target precision, so the median of the best reached precision across all 10 runs has been selected as final result.

\subsubsection{Regression models and their hyper-parameters}

To learn regression models for performance prediction across algorithms, we present results obtained using RF models in STR and MTR learning scenarios. The STR models have been trained with $n\_estimators = 25$ and $max\_depth = 25$, while the MTR model has been trained with  $n\_estimators = 75$ and $max\_depth = 25$. For both scenarios ``\textit{mae}" has been selected to measure the quality of the split. The number of trees in the forest in the MTR scenario has been set to 75, while in STR is set to 25. The maximum depth of the tree is set to 25 in both scenarios. Since in-depth hyper-parameter analysis has not been performed for this study, the number of trees in the MTR RF scenario is actually the sum of trees that appear in the STR RF models.

\subsubsection{50-fold stratified cross validation} To train and evaluate the STR and MTR models, 50-fold cross validation is used based on leave-one instance out. The data set that consists of 1,200 problem instances has been split into 50 folds, where the first fold consists of the first instances from all 24 problems, the second consists of the second instances from all 24 problems, and so on till the 50th fold that consists of the 50th instances from all 24 problems. We repeated the learning process fifty times, where one fold was used for testing, and all the others were used for training the regression model. This evaluation is the most commonly explored scenario in recent studies, since removing all instances from a problem (i.e., leave-one-problem out) does not provide transferable results. This comes from the fact that the COCO benchmark data set was not developed from perspective to cover the diversity required to train ML models.

\subsection{Performance prediction across algorithms}
Here, the results for the performance prediction across algorithms are presented. For this purpose, three RF models in STR learning scenario and one RF model in MTR learning scenario have been trained to predict the original target precision of the three CMA-ES configurations described in Table~\ref{tab:CMA_modules}. The prediction results for other 12 CMA-ES configurations in STR learning scenario are available at our GitHub repository. Looking at the results (see Table~\ref{tab:UC2_models}), it follows that the MTR model provides better results for all three targets, however their difference from the STR models are not practically significant (i.e., small differences in some $\epsilon$ neighbourhood)~\cite{eftimov2019identifying}. For all three configurations, there are benchmark problems for which the STR model provides better results than the MTR model and vice versa. The lowest mean absolute error is obtained for prediction of the target reached by the CMA-ES$_3$ configuration.
\begin{table}[ht]
\centering
\footnotesize
\caption{Mean absolute error for the three CMA-ES configurations across the 50 folds for each COCO benchmark problem obtained by the RF models in STR and MTR scenario.}
\label{tab:UC2_models}
\resizebox{.45\textwidth}{!}{
\begin{tabular}{r|ll|ll|ll}
\hline
\multirow{2}{*}{f\_id} & \multicolumn{2}{c}{CMA-ES$_1$} & \multicolumn{2}{|c}{CMA-ES$_2$} & \multicolumn{2}{|c}{CMA-ES$_3$} \\
  \cline{2-7} 
 & \multicolumn{1}{|l}{STR} & \multicolumn{1}{l}{MTR} & \multicolumn{1}{|l}{STR} & \multicolumn{1}{l}{MTR} & \multicolumn{1}{|l}{STR} & \multicolumn{1}{l}{MTR} \\
\hline
1                    & 0.7950946053                 & \textbf{0.1171230179}        & 0.6274794861                 & \textbf{0.1460143079}        & 0.1813078169                 & \textbf{0.1475319441}        \\
2                    & 104.9619158                  & \textbf{104.750613}          & \textbf{90.64679628}         & 92.39225488                  & 0.3961754563                 & \textbf{0.1962280587}        \\
3                    & 3.762302139                  & \textbf{3.142233438}         & 6.995752656                  & \textbf{3.971825264}         & 5.916780244                  & \textbf{3.657198431}         \\
4                    & \textbf{2.032841432}         & 3.580335038                  & 6.081299376                  & \textbf{4.388368741}         & 4.84942169                   & \textbf{4.44251156}          \\
5                    & \textbf{0.565220947}         & 0.6822621717                 & \textbf{0.381559388}         & 0.4015623417                 & 0.2495819006                 & \textbf{0.09634366455}       \\
6                    & 5.076488528                  & \textbf{1.381378848}         & 3.766669164                  & \textbf{0.4263119407}        & 0.2314116761                 & \textbf{0.2220075017}        \\
7                    & 138.6491104                  & \textbf{137.2151168}         & \textbf{130.4439312}         & 137.0095726                  & 144.0668257                  & \textbf{137.0684437}         \\
8                    & 0.8672745295                 & \textbf{0.7191103104}        & 2.981701837                  & \textbf{0.9494602365}        & \textbf{0.7634002536}        & 0.8021440218                 \\
9                    & 0.7069954314                 & \textbf{0.5819721002}        & \textbf{0.3766386174}        & 0.5615365205                 & \textbf{0.6586060198}        & 0.7178860115                 \\
10                   & 106.7791202                  & \textbf{99.34254356}         & 6.569851731                  & \textbf{4.602317354}         & 2.121300494                  & \textbf{2.005066657}         \\
11                   & 46.43635744                  & \textbf{43.17638024}         & \textbf{5.853058309}         & 9.792925182                  & \textbf{3.024860128}         & 7.555264265                  \\
12                   & \textbf{108.0264528}         & 110.8971307                  & \textbf{7.488099668}         & 7.709678427                  & 3.702828357                  & \textbf{3.299514152}         \\
13                   & \textbf{4.125620833}         & 4.671321592                  & \textbf{0.6456388127}        & 2.061760584                  & \textbf{0.6450998963}        & 1.888585468                  \\
14                   & \textbf{0.6502109504}        & 1.069479504                  & 2.210348191                  & \textbf{0.8706837002}        & \textbf{0.3196071053}        & 0.5746818066                 \\
15                   & 7.491798826                  & \textbf{6.932238256}         & \textbf{6.999607506}         & 7.195588593                  & 7.4197652                    & \textbf{6.891243007}         \\
16                   & \textbf{1.447377783}         & 1.754279299                  & 1.773172292                  & \textbf{1.683888319}         & \textbf{1.098396861}         & 1.397126112                  \\
17                   & 5.953071457                  & \textbf{4.956593959}         & \textbf{3.97947882}          & 5.096178983                  & 6.942321988                  & \textbf{4.954182358}         \\
18                   & \textbf{9.67783419}          & 10.73034895                  & 13.20196118                  & \textbf{9.861722581}         & 12.02241968                  & \textbf{9.765924602}         \\
19                   & \textbf{0.4771408224}        & 1.299007544                  & 1.547377611                  & \textbf{0.8177043829}        & \textbf{0.5974143906}        & 0.8106009347                 \\
20                   & 0.3580273188                 & \textbf{0.2956596327}        & 0.5046414497                 & \textbf{0.3277854333}        & \textbf{0.2253670756}        & 0.2660331833                 \\
21                   & 0.5801459103                 & \textbf{0.5726539085}        & 1.109046526                  & \textbf{0.9361788623}        & 1.346774496                  & \textbf{1.097769283}         \\
22                   & \textbf{1.153179299}         & 1.561458272                  & 2.23296815                   & \textbf{2.019069121}         & 3.284533648                  & \textbf{2.485003216}         \\
23                   & 0.7752912471                 & \textbf{0.5622498049}        & 0.6988707546                 & \textbf{0.5482661113}        & \textbf{0.3432642499}        & 0.6254968227                 \\
24                   & 5.622695422                  & \textbf{4.033250852}         & \textbf{4.158608072}         & 4.553559827                  & \textbf{4.099408527}         & 4.649244225                  \\
\hline
Mean                 & 23.20714868                  & \textbf{22.66769753}                  & 12.55310655                  & \textbf{12.4301756}                   & 8.521119702                  & \textbf{8.150667956}    \\             
\hline
\end{tabular}
}
\end{table}

\begin{figure*}[bht] 
  \begin{subfigure}[b]{0.5\linewidth}
    \centering
    \includegraphics[width=0.75\linewidth]{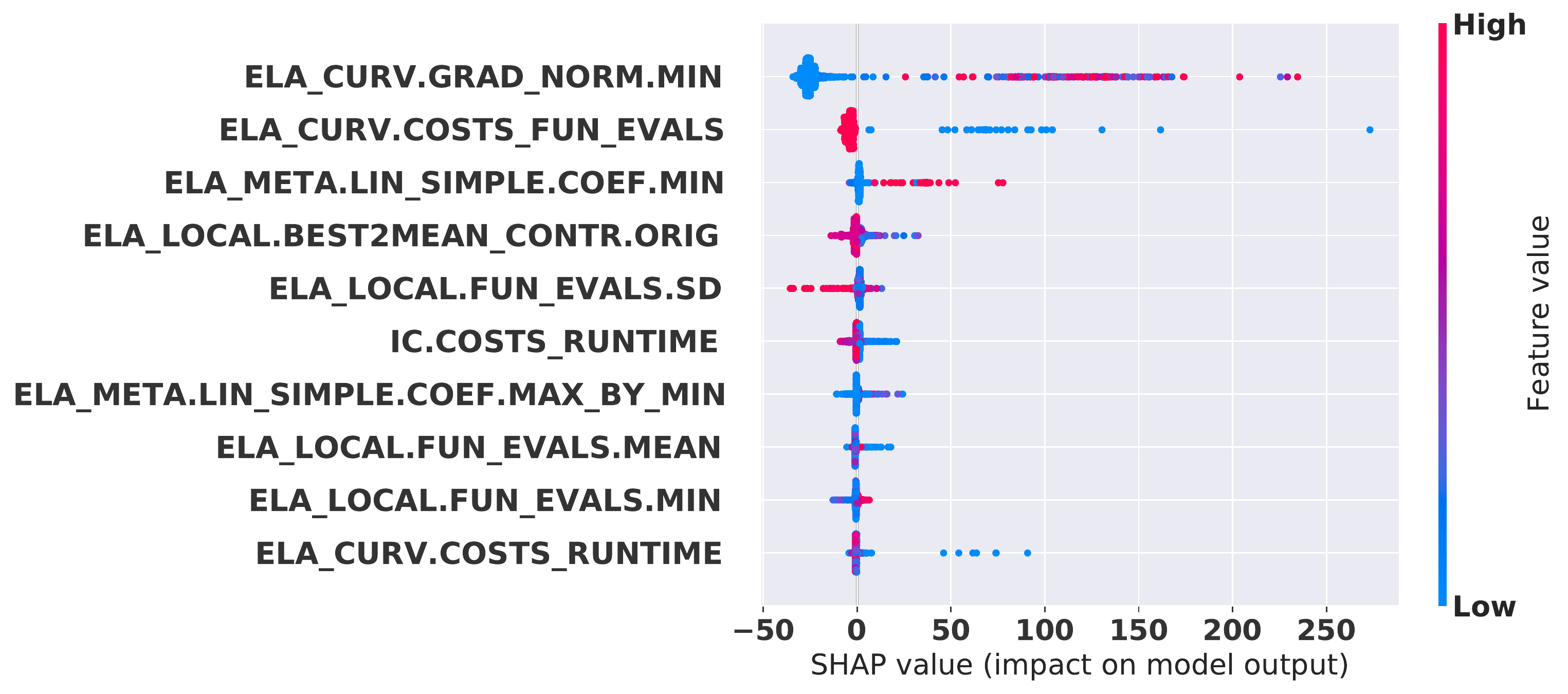} 
    \caption{CMA-ES$_1$} 
    \label{fig11:a} 
    \vspace{4ex}
  \end{subfigure}
  \begin{subfigure}[b]{0.5\linewidth}
    \centering
    \includegraphics[width=0.75\linewidth]{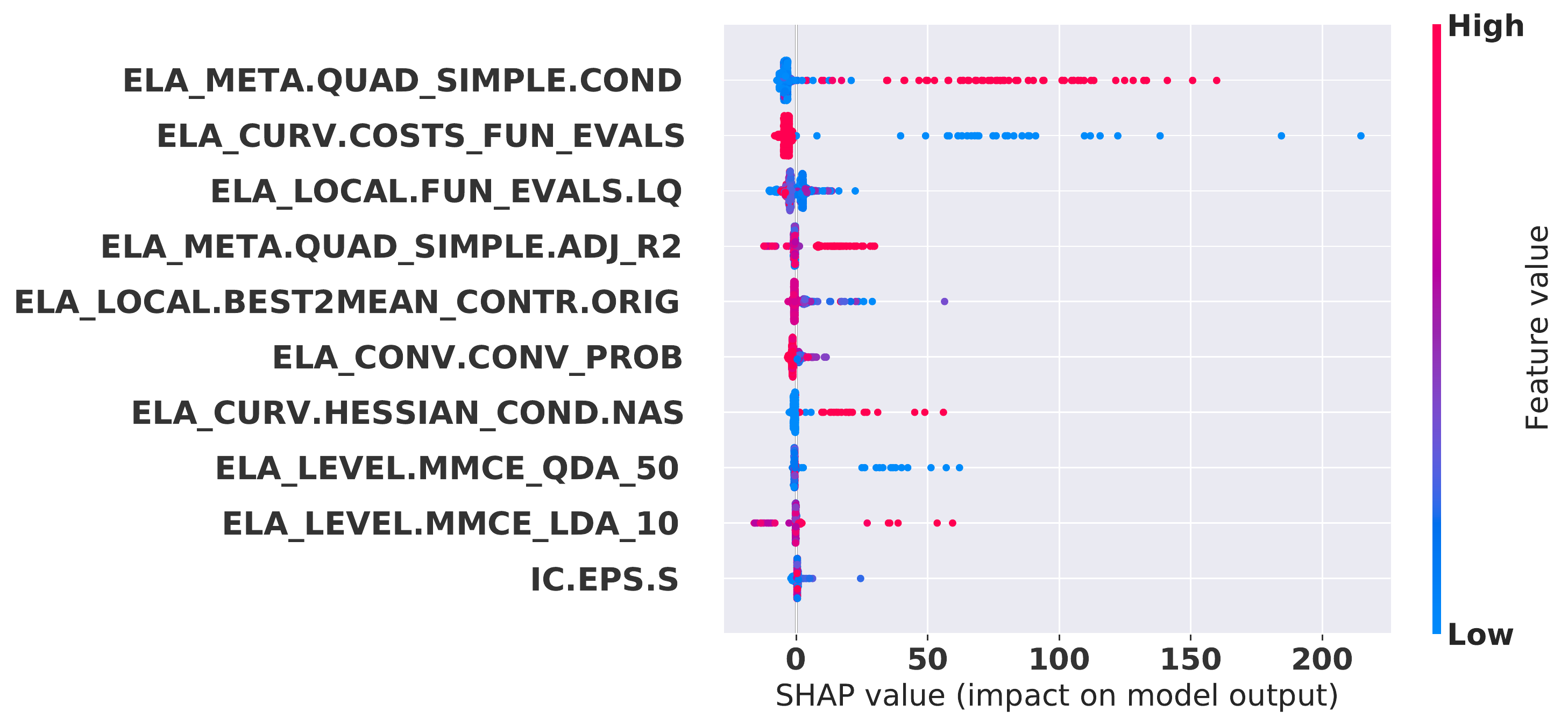} 
    \caption{CMA-ES$_2$} 
    \label{fig11:b} 
    \vspace{4ex}
  \end{subfigure} 
  \begin{subfigure}[b]{0.5\linewidth}
    \centering
    \includegraphics[width=0.75\linewidth]{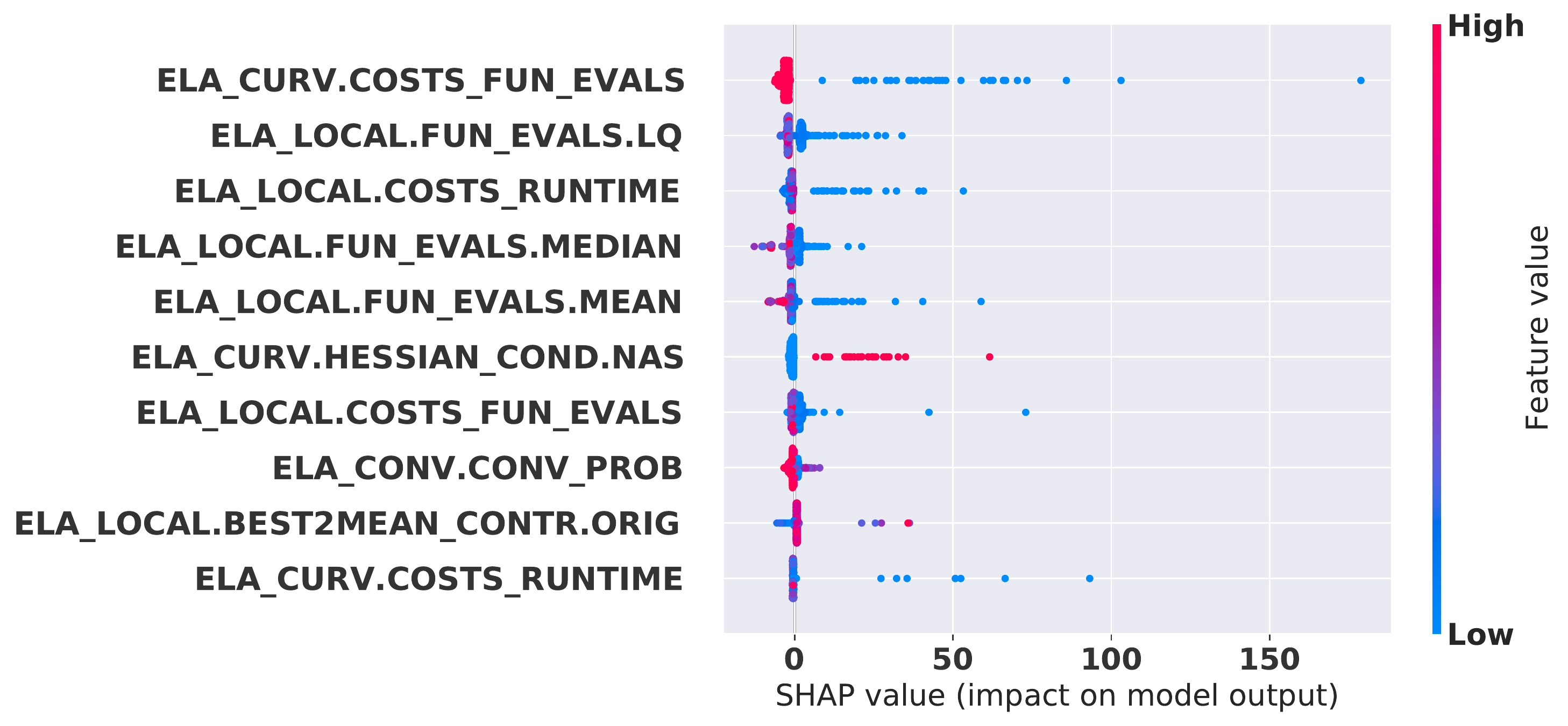} 
    \caption{CMA-ES$_3$} 
    \label{fig11:c} 
  \end{subfigure}
  \caption{SHAP value impact of the STR RF models for predicting the target precision for three CMA-ES configurations.}
  \label{fig:SHAP_UC2_STR} 
\end{figure*}

\subsubsection{Global explanations}
Figure~\ref{fig:SHAP_UC2_STR} presents the SHAP value impact on the STR RF models for prediction of the target precision for the selected CMA-ES configurations. The plots presented in this figure present the positive and negative relationships with the original target precision. The dots presented in the plots correspond to all instances from the training data set (i.e., in our case the first fold). The descending order of the ELA features presents their importance starting from the most important one. The colors used are related to the magnitude of the ELA feature value, where higher values are red and lower value are blue. The impact of the ELA feature value to the target variable prediction is its horizontal location.

From the figure, it follows that the STR RF model for the first configuration shares only four ELA features out of the 10 most important ELA features with the STR RF models for the two others configurations, while the STR RF models for the second and the third configuration overlap in seven ELA features. Looking at the most important features in the STR models, it seems that the difference between different configurations points to the fact that these three different models have different behaviour (i.e., exploration and exploitation powers). This kind of analysis can help us find the most relevant ELA features for each module of the configurable CMA-ES and based on the problem difficulty (its ELA values) the most appropriate modules could be selected. 

 For example, if we have a high value of the ``$ela\_curv.grad\_norm.min$", it contributes by adding a large value to the original target precision reached by the CMA-ES$_1$. This means that high value of this ELA feature is an indication of difficulty in solving the benchmark problem because it takes us away from the optimum reached with adding a large value on the target precision (i.e., error). In addition, looking at the ``$ela\_meta.lin\_simple.coef.min$", lower values do not take us away from the target precision reached  by the CMA-ES$_1$, while higher values of this ELA feature can increase the reached target precision, showing that the benchmark problem is difficult to be solved there. In the future, such kind of analysis can allow us to estimate and rank the problem difficulty concerning the values of the ELA features and their Shapley values. Looking at the most important features it seems that they are all coming from the classical ELA features group.
 
 Figure \ref{fig:SHAP_UC2_STR} presents the explanations obtained from the training data from the first fold for each CMA-ES. To see if these results are consistent across different folds (i.e., if the features importance is consistent within a model across the folds), we have represented each CMA-ES configuration as a vector of 99 Shapley values. For this purpose, we averaged the Shapley value for each ELA feature across all problem instances that belong to each training data fold. The vector with the averaged Shapley values is the CMA-ES configuration representation. 
 \begin{figure}
    \centering
    \includegraphics[scale=0.25]{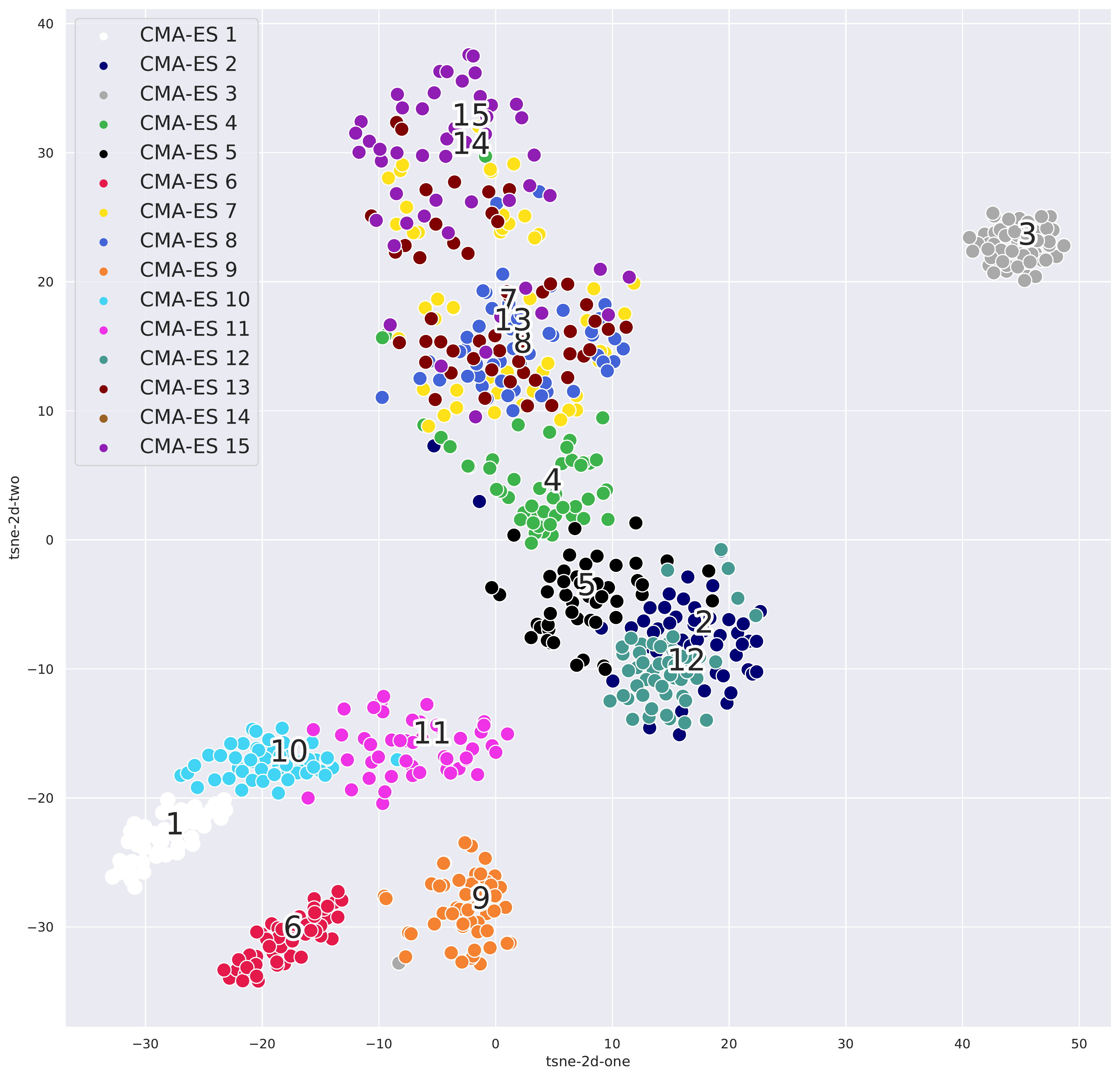}
    \caption{t-SNE visualization of the CMA-ES configurations' representation per each fold. The configurations are represented as vectors of 99 Shapley values (i.e., the importance of each ELA feature to the performance of each configuration, one per each ELA feature).}
   \label{fig:tsne}
\end{figure}

Figure~\ref{fig:tsne} presents t-SNE visualization of each CMA-ES configuration (i.e., three presented in the paper and 12 available at the GitHub repository) in two-dimensional space using their Shapley representation. We have used the default parameters from the t-SNE visualization available from the python package \emph{scikit-learn}~\cite{pedregosa2011scikit}. Looking at the figure, we can assume that the CMA-ES configuration representations are consistent across different folds since their Shapley representations place them close together. This means that to predict the performance of an CMA-ES configuration, the importance of the ELA features utilized by the RF model is the same across the folds. The difference between the CMA-ES configurations points that for different configurations different ELA features are utilized (i.e., their importance change), which points to the fact that they have different exploration and exploitation capabilities. Such kind of representation can help further to analyze algorithms' behaviour and find more similar algorithms, also points to the gaps in the landscape that should be further researched.   

\begin{figure*}[ht] 
  \begin{subfigure}[b]{0.5\linewidth}
    \centering
    \includegraphics[width=0.75\linewidth]{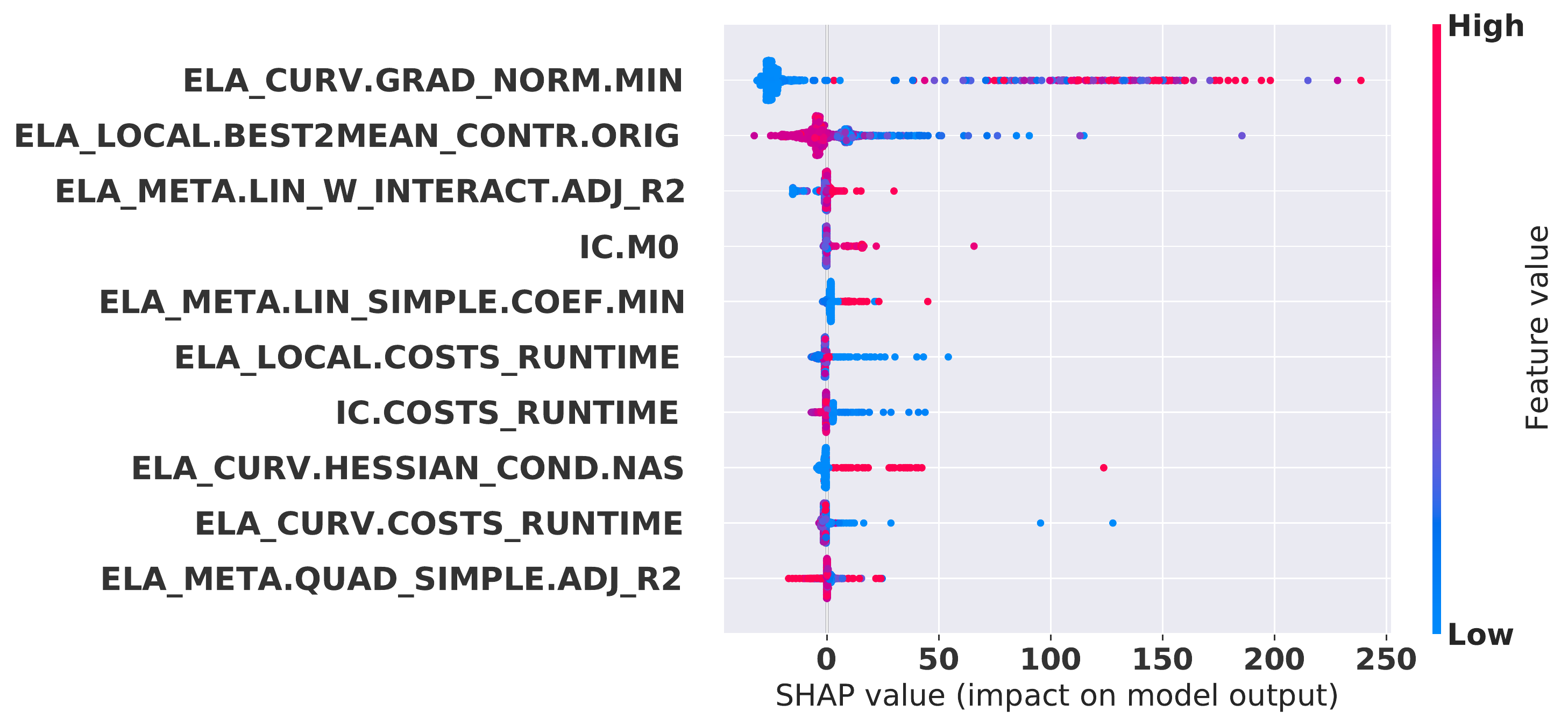} 
    \caption{CMA-ES$_1$} 
    \label{fig12:a} 
    \vspace{4ex}
  \end{subfigure}
  \begin{subfigure}[b]{0.5\linewidth}
    \centering
    \includegraphics[width=0.75\linewidth]{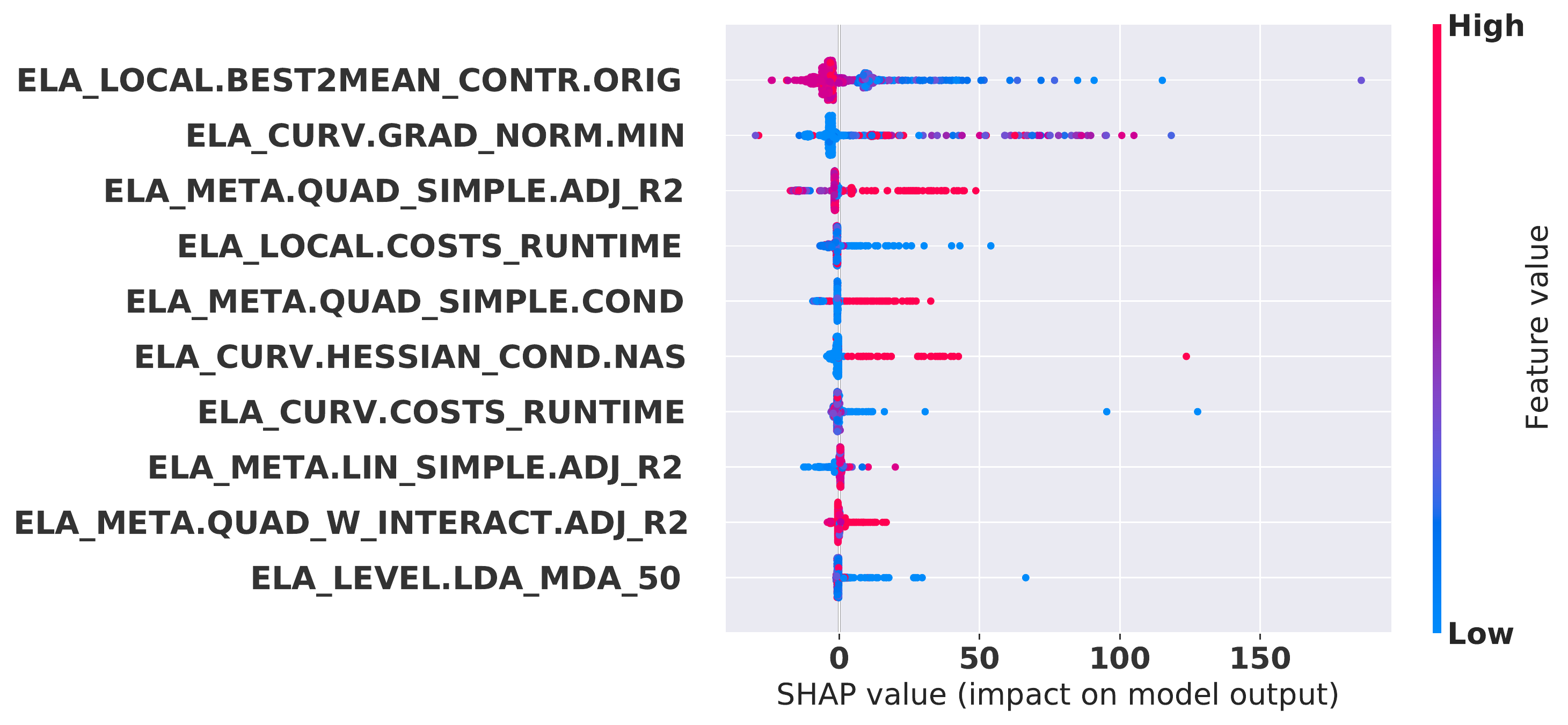} 
    \caption{CMA-ES$_2$} 
    \label{fig12:b} 
    \vspace{4ex}
  \end{subfigure} 
  \begin{subfigure}[b]{0.5\linewidth}
    \centering
    \includegraphics[width=0.75\linewidth]{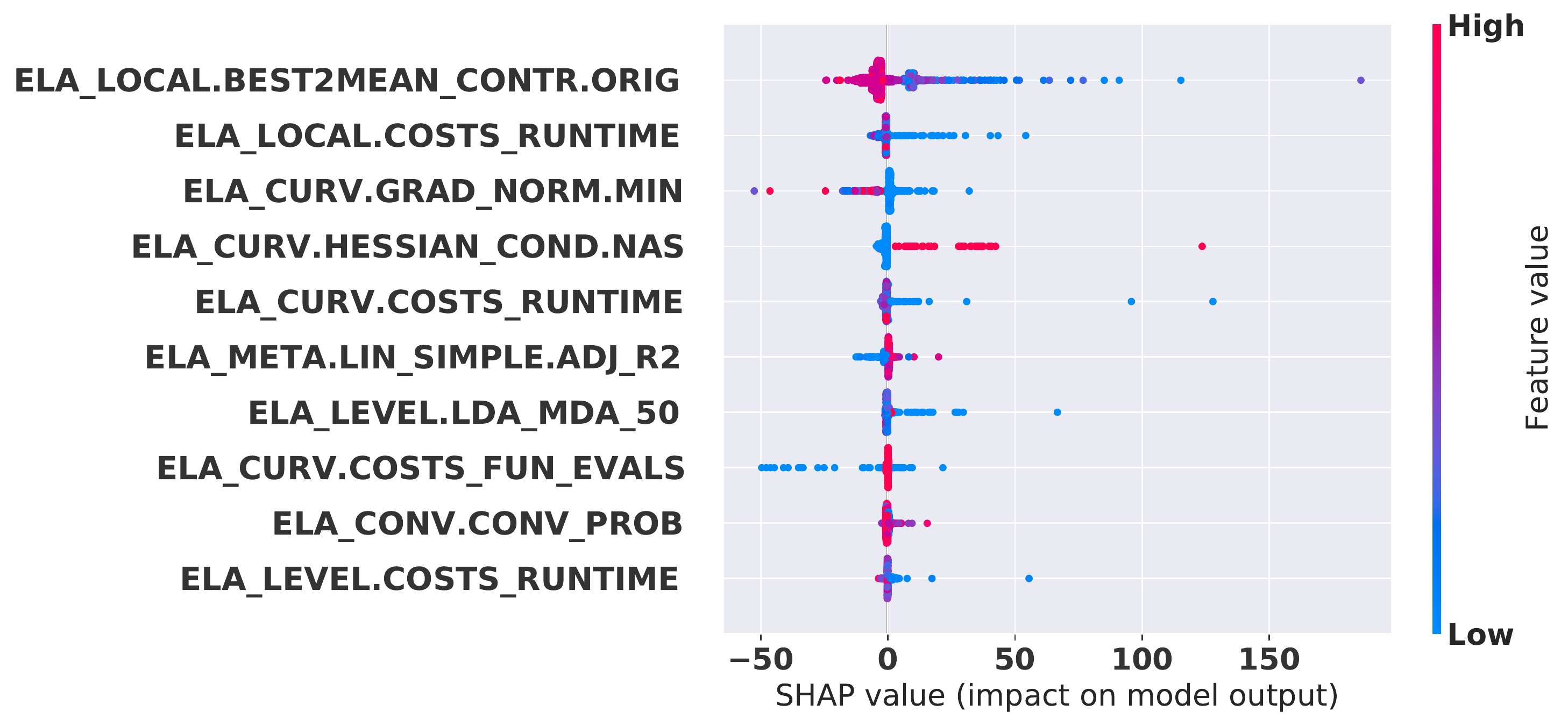} 
    \caption{CMA-ES$_3$} 
    \label{fig12:c} 
  \end{subfigure}
  \caption{SHAP value impact of the MTR RF model for predicting the target precision for three CMA-ES configurations.}
  \label{fig:SHAP_UC2_MTR} 
\end{figure*}

Figure~\ref{fig:SHAP_UC2_MTR} presents the SHAP value impact on the MTR RF model for prediction of the target precision for the selected CMA-ES configurations. From it, it follows that the MTR model overlaps in seven ELA features when it takes decision for the three configurations. It seems that the MTR model tries to find more shared features in favour of the three CMA-ES configurations, which is not a case when STR models are trained.

\begin{figure*}[!bht] 
  \begin{subfigure}[b]{0.5\linewidth}
    \centering
    \includegraphics[width=0.65\linewidth]{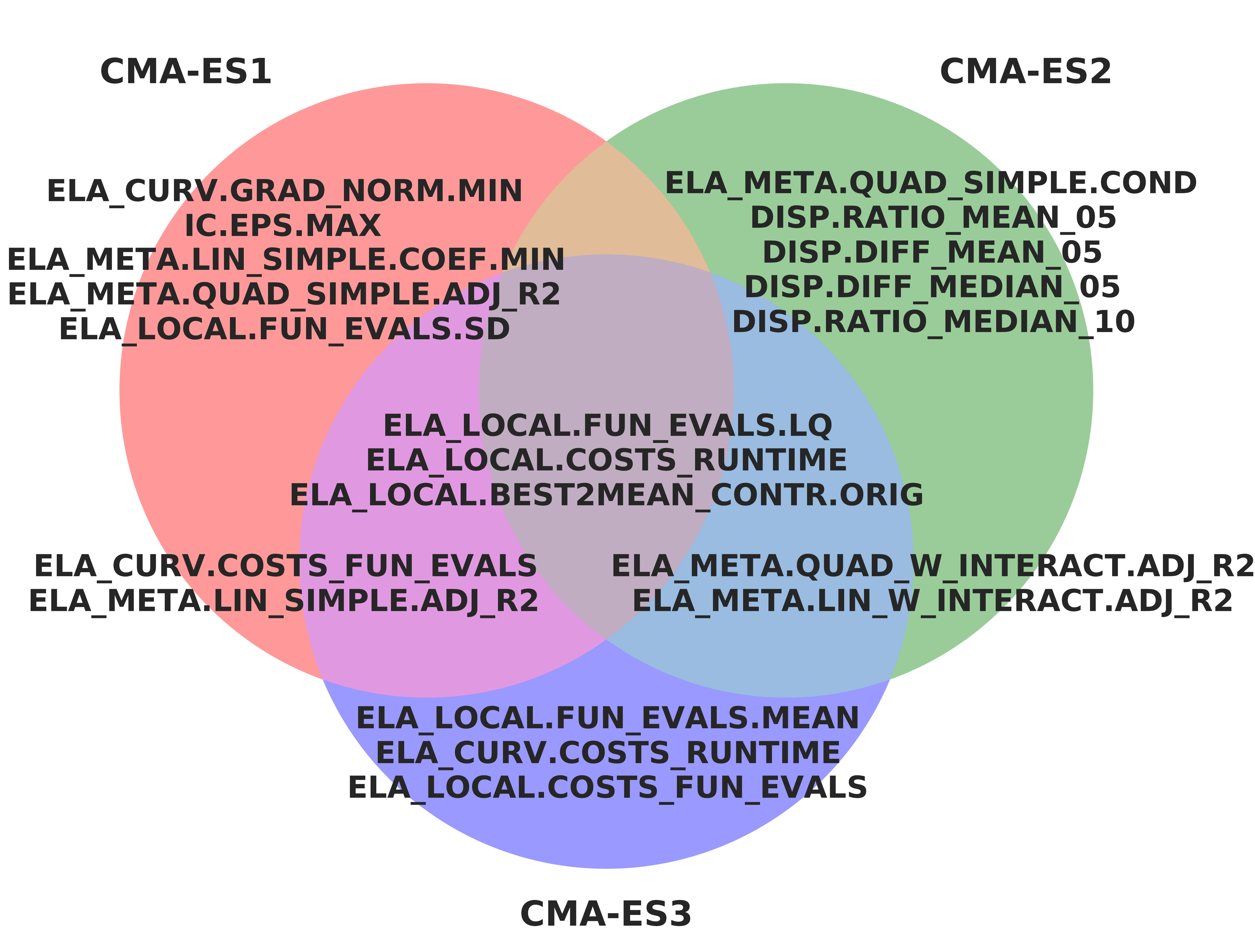} 
      \caption{STR} 
    \label{fig14:a} 
    \vspace{4ex}
  \end{subfigure}
  \begin{subfigure}[b]{0.5\linewidth}
    \centering
    \includegraphics[width=0.65\linewidth]{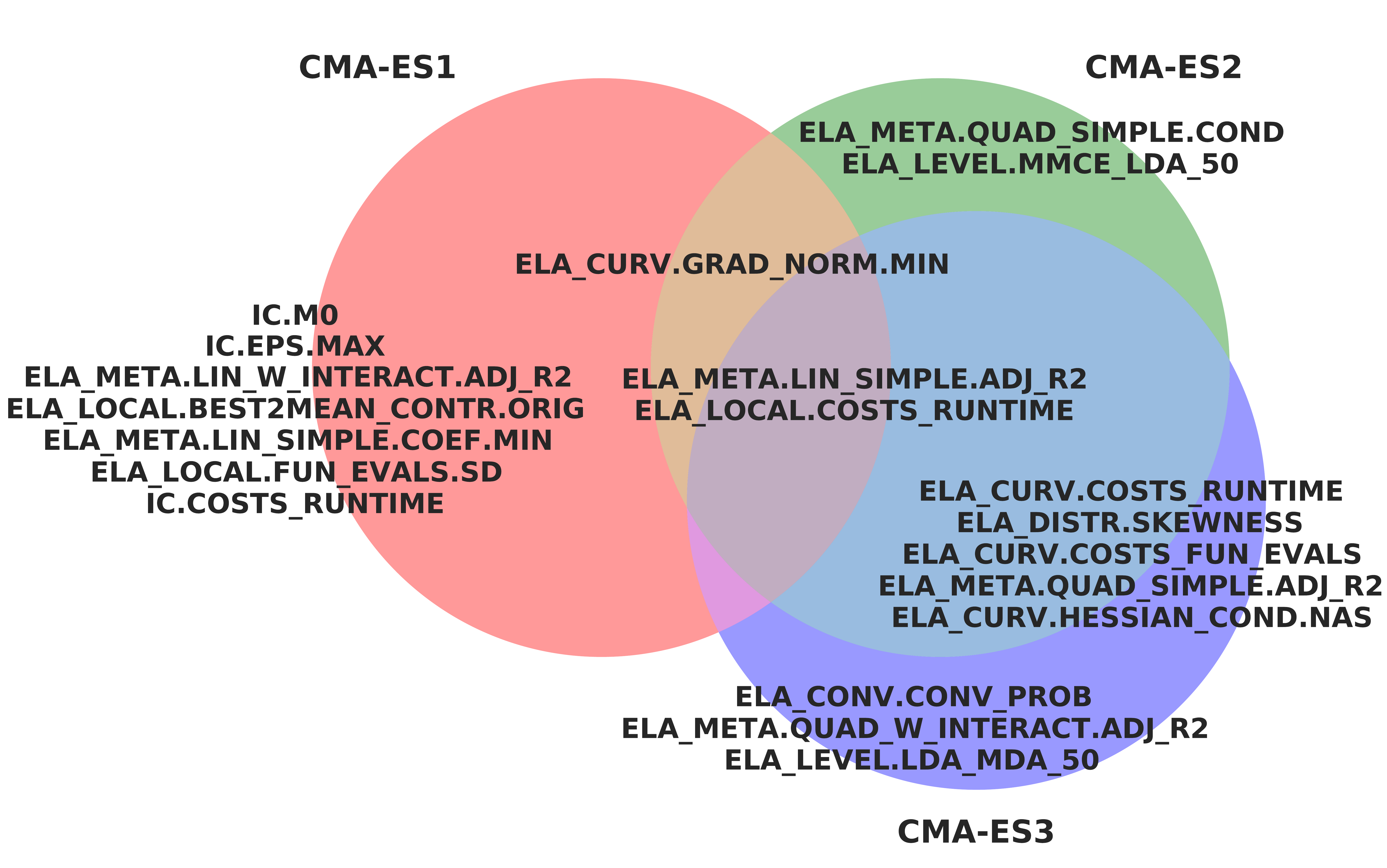} 
    \caption{MTR} 
    \label{fig14:b} 
    \vspace{4ex}
  \end{subfigure} 
   \caption{Intersection between the top 10 most important ELA features utilized by the RF models learn for predicting the target precision of CMA-ES$_1$, CMA-ES$_2$, and CMA-ES$_3$. }
    \label{fig:SHAP_Venn_UC2} 
\end{figure*}

Figure~\ref{fig:SHAP_Venn_UC2} presents the intersection between the top 10 most important ELA features utilized by the RF models learned for predicting the target precision of CMA-ES$_1$, CMA-ES$_2$, and CMA-ES$_3$. To find the top 10 most important ELA features for each modular CMA-ES configuration in STR and MTR scenario, the Shapley results for each ELA features have been first averaged across all problem instances within each training fold, and further across different folds. The most commonly-used features are from the classical \emph{ela} group. We should also point to the fact that for predicting the target precision for the CMA-ES$_2$ in STR scenario, the features from \emph{disp} group seem to be also important. Such kind of representation of the modular CMA-ES configurations can provide us with information for which kind of problems they are suitable.

\subsubsection{Local explanations} 
\begin{figure*}[!ht] 
  \begin{subfigure}[b]{\textwidth}
    \centering
    \includegraphics[width=0.65\textwidth]{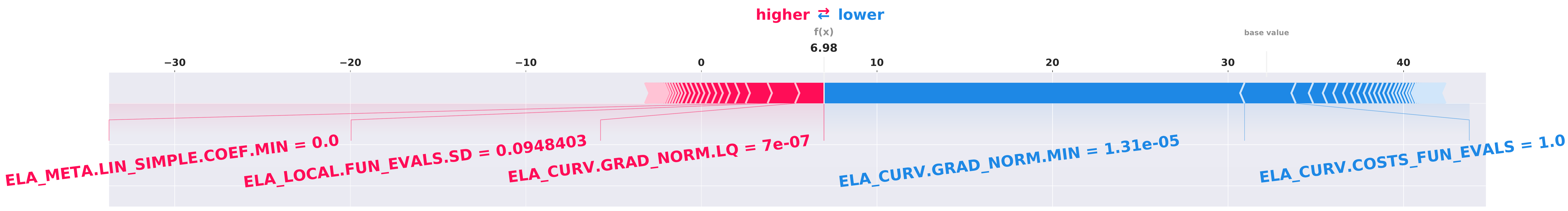} 
      \caption{CMA-ES$_1$} 
    \label{fig15:a} 
    \vspace{4ex}
  \end{subfigure}
  \begin{subfigure}[b]{\textwidth}
    \centering
    \includegraphics[width=0.65\textwidth]{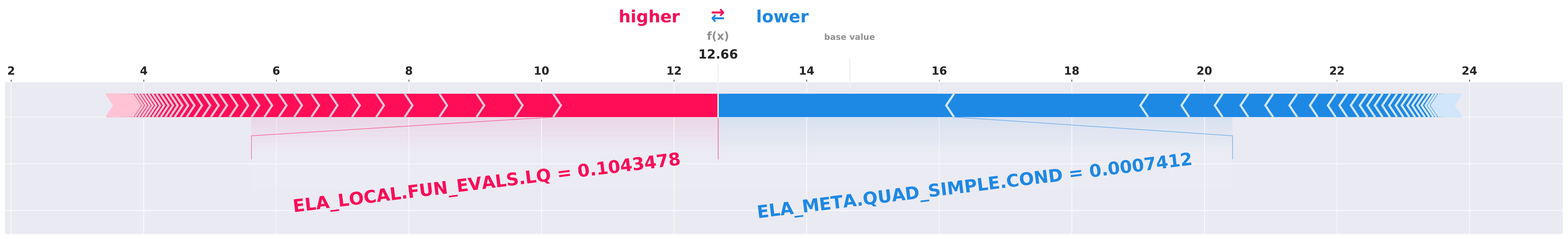} 
   \caption{CMA-ES$_2$} 
    \label{fig15:b} 
 \vspace{4ex}
  \end{subfigure} 
  \begin{subfigure}[b]{\textwidth}
    \centering
    \includegraphics[width=0.65\textwidth]{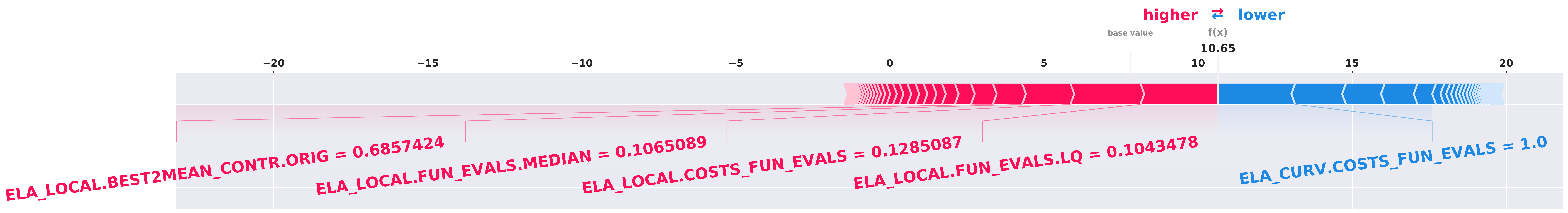} 
   \caption{CMA-ES$_3$} 
    \label{fig15:b} 
 \vspace{4ex}
  \end{subfigure} 
  \caption{SHAP value impact of the STR RF models to explain the predicted target precision reached on the first instance of the fourth benchmark problem.}
  \label{fig:SHAP_local_RF_STR} 
\end{figure*}

\begin{figure*}[!ht] 
  \begin{subfigure}[b]{\textwidth}
    \centering
    \includegraphics[width=0.65\textwidth]{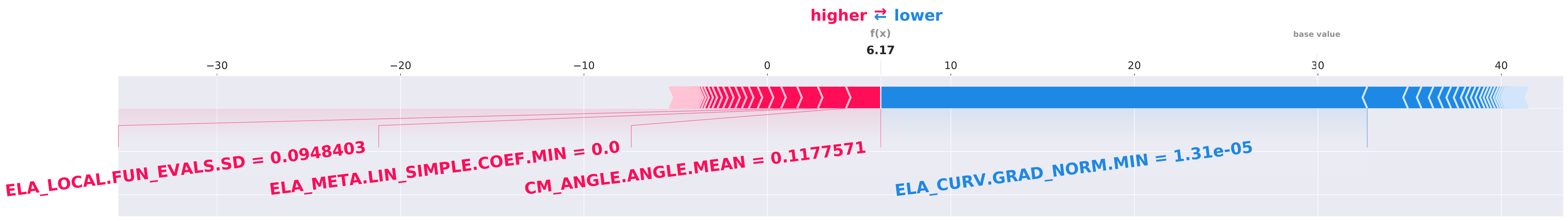} 
      \caption{CMA-ES$_1$} 
    \label{fig15:a} 
    \vspace{4ex}
  \end{subfigure}
  \begin{subfigure}[b]{\textwidth}
    \centering
    \includegraphics[width=0.65\textwidth]{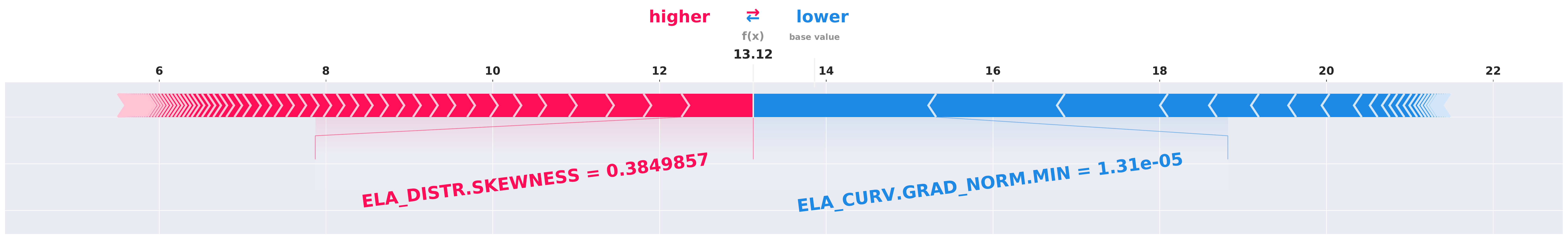} 
   \caption{CMA-ES$_2$} 
    \label{fig15:b} 
 \vspace{4ex}
  \end{subfigure} 
  \begin{subfigure}[b]{\textwidth}
    \centering
    \includegraphics[width=0.65\textwidth]{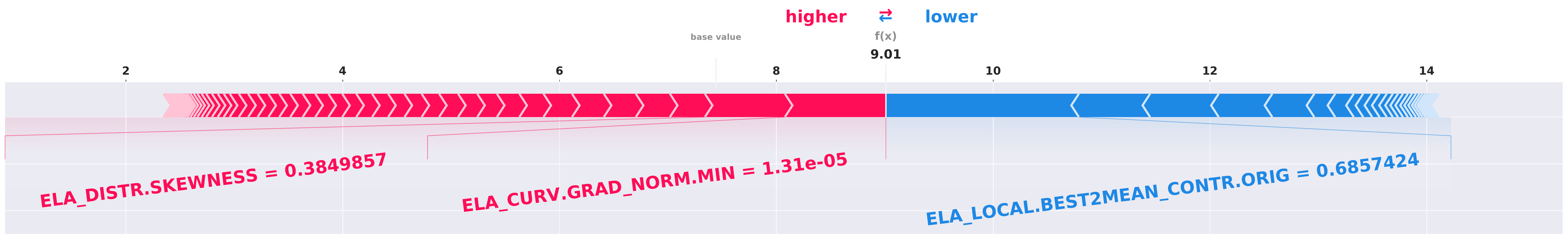} 
   \caption{CMA-ES$_3$} 
    \label{fig15:b} 
 \vspace{4ex}
  \end{subfigure} 
  \caption{SHAP value impact of the MTR RF model to explain the predicted target precision reached on the first instance of the fourth benchmark problem.}
  \label{fig:SHAP_local_RF_MTR} 
\end{figure*}

Figure~\ref{fig:SHAP_local_RF_STR} presents the SHAP value impact of the three STR RF models (one per each CMA-ES configuration) to explain the predicted target precision reached on the first instance of the fourth benchmark problem, while Figure~\ref{fig:SHAP_local_RF_MTR} presents the same results obtained by the MTR RF model. The local explanations provide us the important ELA features that are used for each problem instance separately, which are different from the important global ELA features that are provided using the results across all problem instances. From them, we can see that the second and the third CMA-ES configuration share similar important ELA features that increase the target precision on that problem instance, which are different from the ELA features that increase the precision for the first CMA-ES configuration. This indicates that the second and the third configuration probably have similar behaviour on that problem instance. Due to the page limit, we are not able to provide this analysis for each benchmark problem instances, however the results are available at our GitHub repository.

\section{Conclusions}
\label{sec:conclusions}

Automated algorithm performance prediction plays a crucial part in the automated algorithm selection and configuration tasks. The most common practice from recent studies is to train a supervised ML model using a set of ELA features. However, such models are still black-box with a limited explanations of how each ELA feature of the problem instance influences the end performance result achieved by an optimization algorithm. For this purpose, in this study, we investigated explainable ELA-based regression to predict the algorithm performance. The regression models have been explored in single-target (STR) and multi-target (MTR) scenario for predicting the performance across different algorithms (i.e., target precision reached by different algorithms). The added value is that for each model the explanations are provided on a global (i.e., across all benchmark problem instances) and local level (i.e., for a single problem instance).

Evaluating the approach on the 50 instances from the 24 COCO benchmark problems in combination with three modular CMA-ES configurations by using random forest regression model, it follows that the most important ELA features are coming from the classical \emph{ela} group. The analysis allows us to estimate the contribution of each ELA feature to the performance prediction using a Shapley value. The experimental results showed that different sets of features are important for different problem instances, which points to the fact that further personalization of the landscape space is required when training an automated algorithm performance prediction model.

Next, we are planning to extend this approach on a more diverse algorithm portfolio, which will allow us to represent each optimization algorithm behaviour using the importance of the ELA features that are utilized to predict its performance. This kind of representation can lead to improved algorithm selection and configuration. We are also planning to extend the analysis to different problem dimensions, to analyze patterns that show how features importance behave when the problem dimension increases. In addition, to see the impact of the ML algorithm used for prediction, the explanations will be investigated with some other ML algorithms (e.g., DNN) instead of random forest. Further, we are planning to transfer the importance of the ELA features by selecting different subsets of ELA features per different problem groups. These subsets will be further used to train personalized regression models. The models will use different ELA feature portfolios to predict the optimization algorithm performance for specific problem groups.


\section*{Acknowledgment}

We thank Diederick Vermetten, Leiden University, for providing us the modular CMA-ES performance data. We also thank Pascal Kerschke, University of Dresden, for providing us with the ELA feature values for the 24 COCO functions and their 50 instances. 
Our work was supported by projects from the Slovenian Research Agency (research core funding No. P2-0098, project No. Z2-1867).



\bibliographystyle{IEEEtran}
\bibliography{references}
%

\end{document}